\documentclass[conference]{IEEEtran}
\newcommand{\secref}[1]{Section~\ref{#1}}
\newcommand{\figref}[1]{Figure~\ref{#1}}

\def\ourapproach{$E^2$-CARE}

\usepackage{times}

\usepackage[numbers]{natbib}
\usepackage{multicol}
\usepackage[bookmarks=true]{hyperref}
\usepackage{graphics} 
\usepackage{epsfig} 
\usepackage{times} 
\usepackage{amsmath} 
\usepackage{amssymb}  
\usepackage{tcolorbox}
\usepackage{amsfonts}
\usepackage{multirow}
\usepackage{booktabs}
\usepackage{multicol}
\usepackage{listings}
\usepackage{mathrsfs}
\usepackage[compatibility=false]{caption}

\lstdefinestyle{customstyle}{
    language=TeX, 
    basicstyle=\small\ttfamily,
    keywordstyle=\color{blue},
    commentstyle=\color{green!50!black},
    numbers=left,
    numberstyle=\tiny,
    numbersep=5pt,
    breaklines=true,
    showstringspaces=false,
    frame=none,
}

\lstdefinestyle{LLMQuery}{
  basicstyle=\ttfamily,
  breaklines=true,
  frame=single,
  backgroundcolor=\color{gray!10},
  xleftmargin=0pt,
  columns=fullflexible,
  breakindent=0pt,
  rulecolor=\color{black},
  moredelim=**[is][\textit{}]{\%\%}{\%\%},
  moredelim=**[is][\color{red}\textbf{}\bfseries]{\|\|}{\|\|}
}
\definecolor{ferngreen}{rgb}{0.31, 0.47, 0.26}
\lstdefinestyle{LLMReply}{
  basicstyle=\ttfamily,
  breaklines=true,
  frame=single,
  breakindent=0.2pt,
  backgroundcolor=\color{ferngreen!10},
  xleftmargin=0pt,
  rulecolor=\color{black},
  columns=fullflexible,
}

\lstdefinestyle{customstyle}{
    language=TeX, 
    basicstyle=\small\ttfamily,
    keywordstyle=\color{blue},
    commentstyle=\color{green!50!black},
    numbers=left,
    numberstyle=\tiny,
    numbersep=5pt,
    breaklines=true,
    showstringspaces=false,
    frame=none,
}

\begin{document}

\title{
Embodiment Meets Environment: Toward Context-Aware, Safe Physical Caregiving Robots
}


\author{
  Zhanxin Wu, Ruofei Tong, Jiaying Fang, Tapomayukh Bhattacharjee\\[0.5em]
  Cornell University 
}


%

\twocolumn[{%
    \renewcommand\twocolumn[1][]{#1}%
        \maketitle
        \vspace{-15pt}
 \begin{center}
    \includegraphics[width=\textwidth]{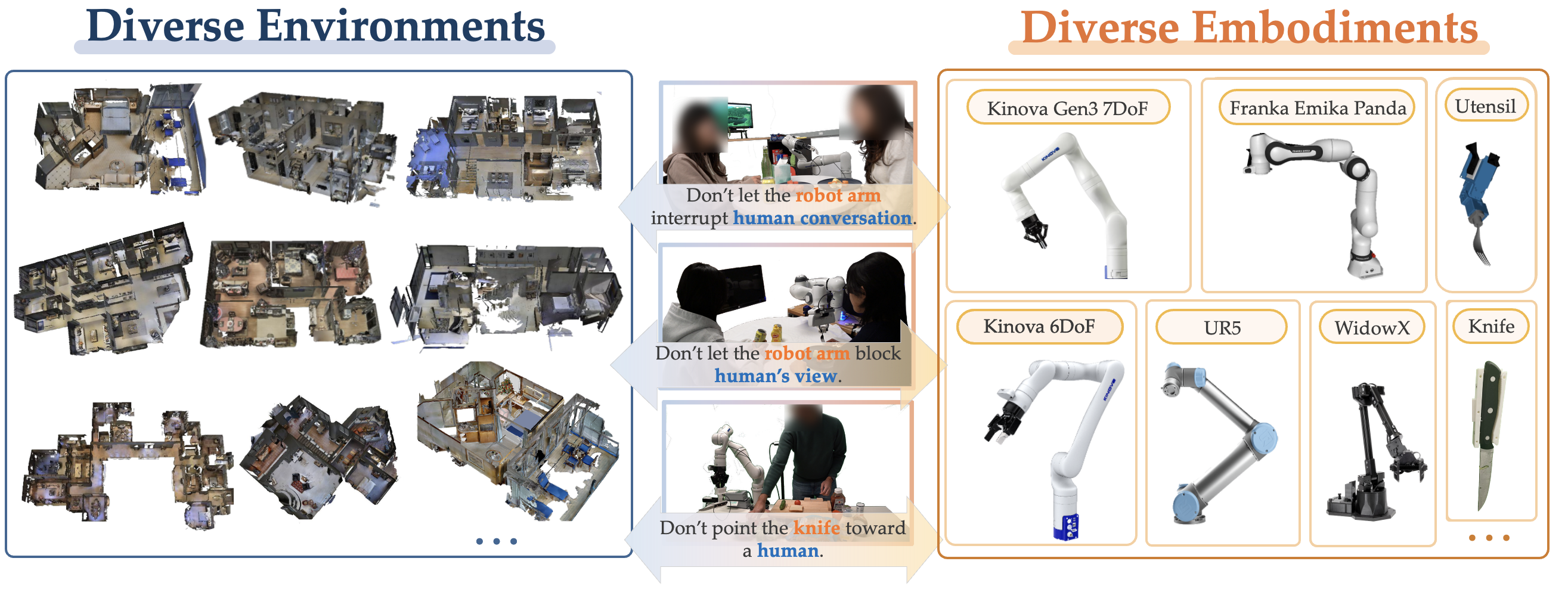}
    \captionsetup{hypcap=false}
    \captionof{figure}{Overview of \ourapproach{}, which enables context-aware caregiving by modeling the joint interaction between the environment and the robot embodiment. Primitive skills are represented as reusable interaction templates and reshaped online via task-specific constraints, allowing safe deployment across diverse environments and robot embodiments.}
    \label{fig:teaser}
    \end{center}
}]

\begin{abstract}
Physical caregiving robots need to assist different users with different tasks in diverse environments, and they come in many embodiments. While substantial progress has been made on individual caregiving tasks, most existing systems remain tightly coupled to specific environments and robot embodiments, and often do not explicitly model or constrain interactions around people, despite humans being special agents in the environment. This motivates a focus on adapting to context that emerges from the joint interaction between the environment and the robot’s embodiment.
We propose \ourapproach{}, a framework that enables context-aware adaptation by representing primitive caregiving skills as interaction templates whose execution is reshaped online. \ourapproach{} represents the environment, the robot, and the human within a unified 3D dynamic scene graph that models these interaction contexts explicitly, and synthesizes task-specific constraints to govern how each skill is executed. By enforcing these constraints at runtime, the same skill templates can be reused zero-shot and safely across diverse environments and robot embodiments. We evaluate \ourapproach{} across four activities of daily living in hundreds of simulated household environments, including assistive home settings, and across diverse robot embodiments, and validate it through user studies on two caregiving tasks with two robots in various real-world environments. Results demonstrate consistent and successful adaptation across these environments and embodiments. Website:
\textcolor{blue}{https://emprise.cs.cornell.edu/e2care}
\end{abstract}

\IEEEpeerreviewmaketitle

\section{Introduction} \label{sec:intro}

Consider a robot assisting people with limited mobility throughout the day: preparing breakfast in a cluttered kitchen, feeding the user at a dining table, and later assisting with grooming beside a wheelchair in a bathroom. Across these activities, robots must help users with different tasks in diverse environments. They may also come in many embodiments, and even change embodiments as they switch tools. In these settings, adaptation is not only about handling new environments or robot embodiments; it is about how the robot’s embodiment interacts with the environment in the human’s presence. While substantial progress has been made on individual physical caregiving tasks, most existing systems remain tightly coupled to specific environments or robot embodiments and require extensive manual reconfiguration as contexts change~\cite{madan2022sparcs, madan2024rabbit, jenamani2024flair, jenamani2025feast}. 
Moreover, many systems do not explicitly model interactions involving people, even though humans are special agents in the environment. Crucially, safe and effective caregiving cannot be achieved by environment understanding or embodiment reasoning alone. For example, when a robot holds a knife near a person, neither “a human is nearby” nor “the robot holds a sharp tool” is sufficient: risk depends on their joint configuration. A sharp end-effector oriented toward a person is dangerous, while the same tool oriented away may be safe. This highlights the need to model context arising from joint interaction between the environment and robot embodiment in the presence of humans.

Existing works have demonstrated the ability to perform a wide range of manipulation tasks and to adapt across environments and robot embodiments, using approaches such as task-and-motion planning~\cite{kumar2024owltamp, kumar2024practice}, modular skill libraries~\cite{wan2024lotus, su2018learning, Skilldiscovery}, and more recently, large-scale learning-based models~\cite{zitkovich2023rt, kimopenvla}. Notable examples include Vision-Language-Action models~\cite{intelligence2025pi05visionlanguageactionmodelopenworld}, which have shown promising performance on tasks such as cleaning kitchens and folding laundry. However, despite these successes, deploying VLA models in real-world, human-centered settings remains challenging. Most existing training datasets are collected in environments without human presence, leaving it unclear how such models generalize to settings that are less structured, involve more dynamic interactions, and impose more stringent safety requirements.

A fundamental limitation of data-driven approaches to physical caregiving is that \textit{the individuals who most require robotic assistance are the least represented in existing datasets}. Collecting large, diverse robot datasets is expensive, time-consuming, and difficult to scale through physical deployment. For example, RT-1~\cite{rt12022arxiv} required approximately 130,000 demonstrations collected over 17 months, and achieving dataset sizes comparable to foundation models in other domains (e.g., large language models) would require millions of robot-hours~\cite{khazatsky2024droid, 2018sergey_handeye, fang2024rh20t, open_x_embodiment_rt_x_2023}. This challenge is further amplified in physical caregiving scenarios, which inherently require large-scale human-robot interaction data. Caregiving tasks such as feeding and dressing involve continuous interaction with humans, making data collection risky and difficult to scale. While a small number of datasets capture how human caregivers perform ADLs~\cite{liang2025openrobocare}, none capture the full complexity of physical human-robot interaction. As a result, developing foundation models for physical caregiving faces challenges.

Our key insight is that robust caregiving behavior cannot be specified by the environment or the robot embodiment alone, but must be grounded in context that arises from how a particular embodiment interacts with a particular environment in the presence of humans. As a result, caregiving skills should not be treated as fixed policies, but as adaptable interaction templates whose execution depends on the current environment-embodiment context. Building on this insight, we propose \ourapproach{}, a framework that explicitly represents the environment, the robot embodiment, and the human within a unified 3D dynamic scene graph. Given this representation of contexts, \ourapproach{} synthesizes constraints that shape how each skill is carried out at runtime. For example: do not point the knife blade toward a human. We distinguish between hard constraints, which encode safety-critical conditions that must always be satisfied to guarantee human safety, and soft constraints, which capture task-specific requirements and are satisfied whenever possible. We enforce hard constraints and maximize satisfaction of soft constraints during execution using control barrier functions, enabling the same skill templates to be safely reused across diverse environments and robot embodiments without retraining.

We evaluate \ourapproach{} across four activities of daily living (ADLs) in hundreds of simulated household environments with diverse robot embodiments, and further validate it through user studies on two real-world caregiving tasks using two robot embodiments. Our results demonstrate that explicit reasoning over environment-embodiment interaction is critical for safe and effective adaptation in physical caregiving scenarios.

\section{Related Work}

 \begin{figure*}[ht]
     \centering
     \includegraphics[width=0.9\linewidth]{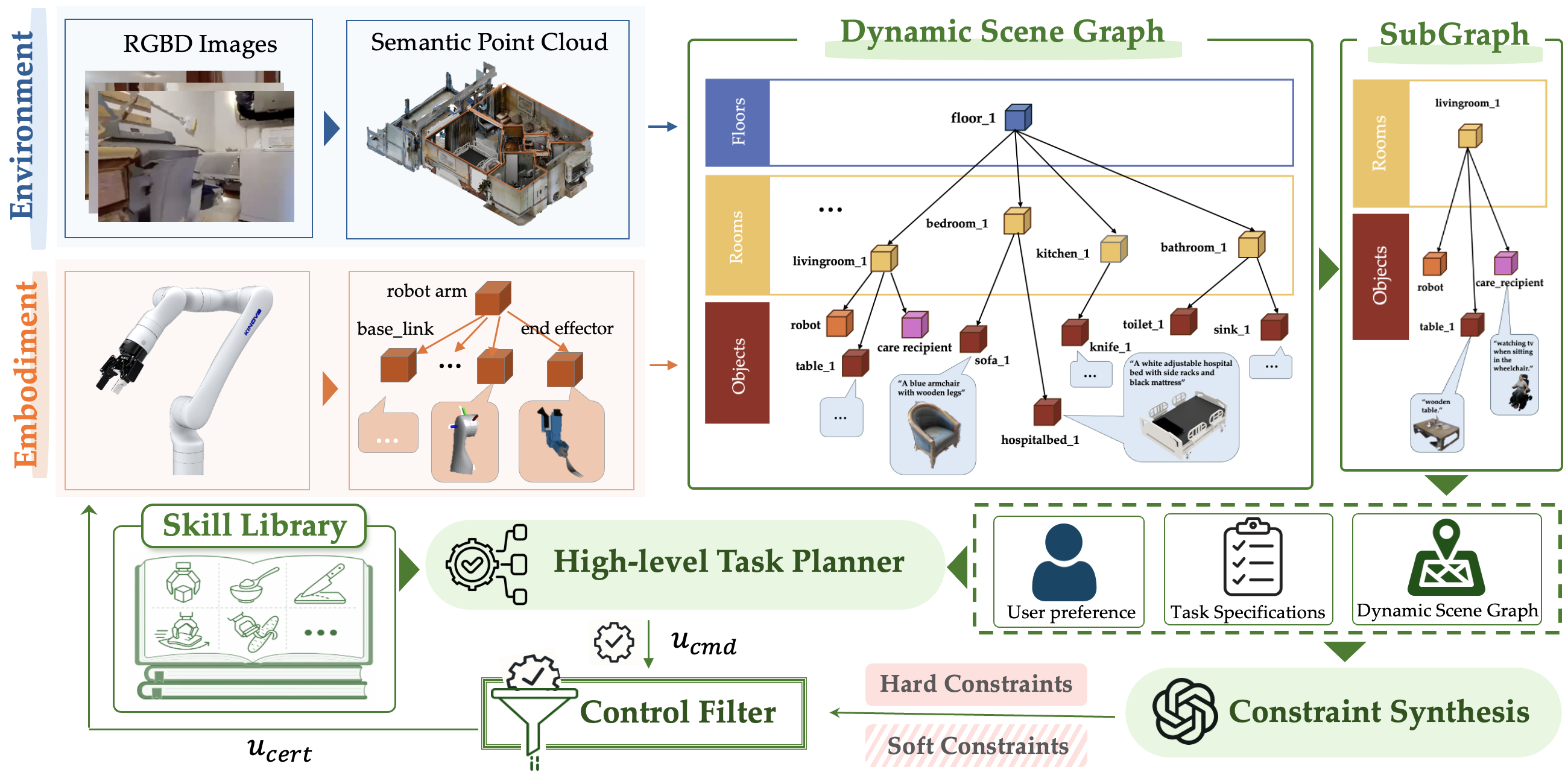}
     \caption{$E^2$-CARE framework. We represent the environment, the robot, and the human within a unified 3D dynamic scene graph that models these interaction contexts explicitly, and synthesizes task-specific constraints to govern how each skill is executed. By enforcing these constraints at runtime, the same skill templates can be reused zero-shot and safely across diverse environments and robot embodiments.}
     \label{fig:framework}
     \vspace{-15pt}
 \end{figure*}

\textbf{Environmental Adaptation of Physical Caregiving Robots.} 
Prior work has developed physical caregiving robots for a variety of activities of daily living (ADLs), including feeding~\cite{jenamani2024flair, hri2025feeding, gallenberger2019transfer}, grooming~\cite{zhao2025dymohair, iros2021inproceedings}, dressing~\cite{sun2024dress, zhang2022learningdressing, cleg2020dressing, iros2025dressing}, bathing~\cite{zlatintsi2018multimodal, madan2024rabbit, gu2024vttb, liu2024skingrip}, and meal preparation~\cite{Dong2021Peeling}\cite{MORPHeus2024Peeling}. These systems typically achieve reliable performance by designing task-specific pipelines that are often tightly coupled to a particular environment~\cite{jenamani2025feast, madan2024rabbit, yoo2025moehair}. These assumptions fundamentally limit generalization. When deployed in real-world household environments with different room geometries, furniture types, object arrangements, or human presence, these systems often require extensive manual reconfiguration~\cite{hri2025feeding, jenamani2025feast, jenamani2024flair}. As a result, existing physical caregiving robots struggle to adapt zero-shot to the physical and social characteristics of unseen household environments. In contrast, we explicitly reason about environmental structure where humans are treated as special agents, allowing caregiving skills to adapt at execution time rather than relying on environment-specific pipelines.

\textbf{Embodiment Adaptation of Physical Caregiving Robots.}
Existing work has demonstrated physical caregiving across multiple ADLs on a variety of robot embodiments~\cite{park2020active, grannen2022learning, PARK2020103344, obi, neater, wu2025savor, palacin2019extending}. However, most systems are designed around a single robot embodiment and do not explicitly model how embodiment differences affect task execution~\cite{jenamani2024flair}. Consequently, when a system is deployed on a different robot, manual modification is often required. In these systems, tools used for caregiving are typically treated as interchangeable end effectors rather than as components that fundamentally alter the robot’s effective embodiment and its interaction constraints (e.g., sharp tools near humans or a wet sponge near electronics). As a result, embodiment-dependent execution constraints are rarely modeled explicitly and are instead encoded implicitly through task-specific design choices. Beyond technical system design, prior work has shown that robot embodiment strongly influences user acceptance in caregiving contexts, and that assistance needs evolve over time as users age~\cite{app15020891, tsui2024perspective, kemp2022design}. Together, these findings highlight that embodiment is central to both physical interaction and long-term caregiving deployment. Our work explicitly models robot embodiment and its interaction with the environment and humans, enabling skills to be reused across embodiments.

\textbf{Adaptation to Environment and Embodiment in Robot Manipulation.} 
Prior work addresses both environmental and embodiment variation by training generalist manipulation policies on large, diverse datasets spanning multiple scenes and robot embodiments~\cite{black2024pi0Vla, kimopenvla, intelligence2025pi05visionlanguageactionmodelopenworld, zitkovich2023rt}. While promising, these data-driven methods typically rely on large-scale supervision and require substantial amounts of data and computation~\cite{li2024cogact, chen2024mirage, yang2024pushing, fan2025interleavevlaenhancingrobotmanipulation}. Moreover, environment and embodiment are often handled implicitly via data coverage, rather than being explicitly modeled. 
Other work explicitly represents aspects of the environment to support task execution~\cite{semantic-manipulation, ravichandran_roboguard}. 
However, these approaches typically focus on static environments and consider only a limited set of objects, either ignoring human presence or predefining fixed safety distances from humans. Consequently, they remain insufficient for physical caregiving scenarios, where robots must continuously operate in close proximity to humans. Our work addresses this gap by explicitly reasoning about environment-embodiment interactions in the presence of a human, conditioned on the task, enabling existing caregiving skills to be modulated at execution time without policy retraining or dataset expansion.

\section{Problem Formulation}

We consider the problem of \textbf{embodiment- and environment-aware physical caregiving robots} performing \emph{activities of daily living} (ADLs) in diverse human living environments. The robot operates with an embodiment $\mathcal{R}$. The embodiment may change over time as the robot equips or removes tools (e.g., a utensil for feeding or a knife for meal preparation). We assume $\mathcal{R}$ is specified by a Unified Robot Description Format (URDF), defining the robot’s kinematics, degrees of freedom, and tool attachments. The robot executes tasks within an environment $\mathcal{E}$ that varies in spatial layout, furniture geometry, object configurations, and human presence. We assume access to the primitive skill library $\mathcal{L} =\{l_1, l_2, ...l_k\}$, consisting of parameterized skills (e.g., \emph{navigate\_to}$(p)$ for $p \in \mathbb{R}^3$) and learned skills (e.g., \emph{cutting}), which can be composed to perform ADL tasks. The central challenge is to execute caregiving skills for activities of daily living safely and effectively across variations in both $\mathcal{R}$ and $\mathcal{E}$. Given an ADL task $\tau$ and a user preference expressed in natural language, the robot receives an observation $o_t$ at timestep $t$, where
$o_t = (i_t, f_t, p_t)$, with
$i_t \in \mathbb{R}^{W \times H \times 4}$ an RGB-D image,
$f_t \in \mathbb{R}^6$ the force-torque readings ($\mathbf{0}_6$ if unavailable), and
$p_t$ the joint state of the robot.
At timestep $t$, the robot selects a primitive skill $l_t \in L$ and executes an action $a_t \in l_t$.

\section{\ourapproach{}}
\label{sec:Method}
We present \ourapproach{}, a framework for environment-embodiment-aware physical caregiving robots (\figref{fig:framework}). \ourapproach{} models the environment, the robot, and the human within a unified 3D dynamic scene graph to explicitly represent their geometric and semantic interactions from sensory observations. Given a task $\tau$ and a skill library $\mathcal{L}$, a Planning Domain Definition Language (PDDL)-based planner generates a sequence of primitive skills using symbolic predicates grounded in the scene graph. Conditioned on the task $\tau$ and user preference, an LLM-based reasoning module performs semantic reasoning to analyze the scene graph and derive embodiment- and environment-dependent interaction constraints for each skill, including hard safety constraints and soft task preferences. These constraints are converted into state-dependent control conditions, where hard constraints are enforced at execution time through a quadratic-program (QP) control filter based on control barrier functions (CBFs), while soft constraints guide optimization when feasible. This design enables the same skill primitives to be reused safely and efficiently across diverse environments and robot embodiments by reshaping their execution through context-dependent constraints. We detail each component below.

\subsection{3D Dynamic Scene Graph Generation}
3D Dynamic Scene Graph $G(V_t,E_t)$ at time step $t$ is a layered, hierarchical representation that abstracts dense 3D reconstructions into higher-level spatial concepts such as objects, agents (including the robot and humans), and rooms, while explicitly modeling their spatio-temporal relationships (e.g., “object A is in room B at time $t$”). This structure enables the system to reason over embodiment-environment interactions using semantic entities rather than raw geometry alone.

Each node $i \in V_t$ corresponds to a spatial entity (e.g., a manipulable object, a semantic region, the robot, or a human) and stores attributes $(p_i, f_i)$, where $p_i$ is a geometric representation (e.g., a point cloud) and $f_i$ is a semantic descriptor. Edges encode pairwise spatial and semantic relations such as adjacency, containment, and contact. Nodes are updated over time through multi-frame association and tracking, allowing the graph to maintain persistent identities and support dynamic reasoning as the scene evolves. 

\paragraph{Environment}
We construct the environment graph using a SLAM pipeline~\cite{loo2025open, Rosinol20rss-dynamicSceneGraphs, gu2024conceptgraphs}. At each discrete time step, the robot receives an observation $(i_t, u_t)$, where $i_t \in \mathbb{R}^{W \times H \times 4}$ is an RGB-D image and $u_t \in \mathbb{R}^6$ denotes IMU measurements, with the camera pose estimated by the SLAM backend. The backend maintains a metric map that is abstracted into semantic entities and relations. For each frame $i_t$, open-vocabulary object detection (RAM~\cite{zhang2023recognize}, GroundingDINO~\cite{liu2023grounding}) and segmentation (SAM~\cite{kirillov2023segany}) produce a set of object masks $\{m_k^t\}$, where each mask corresponds to a detected object k. These masks are associated across frames and projected into 3D point clouds $\{p_k^t\}$ using depth and pose estimates. The resulting graph maintains a hierarchical structure: at a high level it provides a compact semantic abstraction of the environment, while at a lower level it grounds semantics into geometry, yielding a representation that supports downstream planning and execution-time reasoning.

\paragraph{Embodiment} 
We convert the robot’s URDF into a kinematic graph that captures the structural organization of its components and add it into the 3D dynamic scene graph. Rather than treating the robot as a single rigid entity, we represent it as a set of functional components, including the mobile base, arm, and end effector, where the end effector may change as the robot equips different tools. At the geometric level, each component is associated with its mesh model and a collision proxy representation. Following~\cite{morton2025oscbf}, we approximate each link using a set of collision spheres anchored to the kinematic structure, enabling efficient computation of collision checking during execution. The robot embodiment is represented hierarchically: at a high level, it provides a functional abstraction of the robot (mobile base, arm, end effector); at a lower level, it encodes link-level geometry, yielding a grounded representation that supports robot behavior adaptation.

\subsection{High-Level Planning}
Given a task $\tau$, a skill library $L$ (see Appendix), and the current 3D dynamic scene graph $G(V_t,E_t)$, we first query an LLM to identify task-relevant entities and prune irrelevant nodes, producing a task-conditioned subgraph $G(V_t’,E_t’)$. For example, during a feeding task, nodes corresponding to unrelated regions (e.g., a bathroom) are removed. This subgraph preserves the semantic and geometric structure necessary for the task and serves as the environment-embodiment context while reducing reasoning complexity. We then ground the task-relevant subgraph $G(V_t’,E_t’)$ into a symbolic PDDL state by mapping nodes to objects, and perform high-level planning using a PDDL-based planner to generate a task plan that achieves the goal.

\begin{lstlisting}[basicstyle=\ttfamily\small]
# an example skill primitive 
(:action Pickup
  :parameters (?obj - object)
  :precondition (and (GripperFree)
                     (Reachable ?obj))
  :effect (and (Holding ?obj)
               (not (GripperFree))
)
\end{lstlisting}

\subsection{Constraint Synthesis}
Given a PDDL-planner skill sequence, \ourapproach{} treats each primitive skill as an interaction template and adapts them at execution time by synthesizing environment-embodiment interaction constraints for each skill. These constraints specify how a skill should be executed safely and appropriately in the current context, taking into account both environmental conditions and the effective embodiment of the robot.

We define two types of constraints: hard constraints and soft constraints. Hard constraints encode safety-critical states and must never be violated, such as ensuring that the robot does not collide with a human. These constraints are predefined and enforced deterministically using control barrier functions. Soft constraints encode user preferences that affect execution behavior but may be relaxed when necessary, such as avoid occluding the user’s view. 
\paragraph{Constraint Representation}

We represent constraints using a symbolic interface that separates
\emph{what is measured} from \emph{what rule is enforced}.
For example, a safety constraint such as
“keep the end-effector 30\,cm away from a human”
measures distance and enforces a minimum bound.
This structure allows semantic instructions to be compiled into control-safe rules. Formally, each constraint is encoded as a tuple
\begin{equation}
c_i = (s_i, o_i, r_i, m_i, \theta_i),
\end{equation}
where $s_i$ is a robot embodiment entity (e.g., an end-effector or arm link) from $G(V'_t, E'_t)$
and $o_i$ is a scene entity from $G(V'_t, E'_t)$
(e.g., a human or object).
The relation type $r_i \in \mathcal{R}$ specifies the geometric quantity
being measured, and $\mathcal{R}$ is defined as 
\begin{equation}
\mathcal{R} =
\{\text{distance}, \text{region}, \text{orientation}, \text{velocity}\}.
\end{equation}
for all tasks. 
The mode $m_i$ specifies the behavioral rule applied to that measurement,
such as enforcing a lower bound, an upper bound, or directional avoidance.
The parameter vector $\theta_i$ contains numeric thresholds
(e.g., distance limits or angular bounds).
Together, $(r_i, m_i, \theta_i)$ define both the measured quantity
and the rule governing it.
Every constraint therefore corresponds to a safe set in state space
that can be compiled into a differentiable control barrier function.

\paragraph{Hard Constraints}
Hard constraints encode safety-critical conditions that must never be violated, such as collision avoidance, joint limits, and minimum separation between the robot and humans. 
These constraints are predefined to guarantee safety.

\begin{tcolorbox}[colback=white, colframe=black,
title={\textbf{Example: Cutting with a knife}},
fonttitle=\bfseries,
boxsep=0.5mm, left=1mm, right=1mm, top=0.5mm, bottom=0.5mm, arc=0mm]
\linespread{0.9}\selectfont
Joint limits: \\
$(\text{robot}, \varnothing, \text{joint}, \text{inside-limit}, \text{joint\_limit\_bounds})$ \\[2pt]
Self-collision avoidance: \\
$(\text{robot link}, \{\text{other links}\}, \text{distance}, \text{keep-distance}, 0.05)$ \\[2pt]
Human safety distance: \\
$(\text{robot}, \{\text{human}\}, \text{distance}, \text{keep-distance}, 0.1)$
\end{tcolorbox}

In general, we define $x$ as the current state of the robot, the human, and objects from the dynamic scene graph. A constraint is represented as a control barrier function $h(x)$: $h(x) \ge 0$ indicates safe states, while $h(x) < 0$ corresponds to unsafe states. For example, a human-robot distance constraint can be written as $h(x) = \text{dist}(\text{robot}, \text{human}) - d_{\min}$ where $d_{\min}$ is the minimum distance threshold. Formally, each hard constraint is represented as a control barrier function
$h_{\text{hard}}(x) \ge 0$.

\paragraph{Soft Constraints}
Soft constraints encode semantic, social, or user preferences that should be
satisfied when feasible but may be relaxed to preserve safety or feasibility.
For example, the robot should avoid blocking a user’s view while they are watching TV.

\begin{tcolorbox}[colback=white, colframe=black,
title={\textbf{Example: Feeding while the user is watching TV}},
fonttitle=\bfseries,
boxsep=0.5mm, left=1mm, right=1mm, top=0.5mm, bottom=0.5mm, arc=0mm]
\linespread{0.9}\selectfont
$(\text{robot}, \{\text{human, tv}\}, \text{region}, \text{outside-region}, 0.1)$ \\
$(\text{robot}, \varnothing, \text{velocity}, \text{limit-speed}, 0.1)$
\end{tcolorbox}

Soft constraints are generated by querying an LLM-based reasoner with the
task specification, user preferences, and the task-relevant subgraph
$G(V_t',E_t')$.
The reasoner outputs symbolic constraint tuples, which are grounded using
geometric attributes stored in the scene graph.
Each grounded constraint is compiled into a differentiable barrier function
\begin{equation}
h_{\text{soft}_i}(x) \ge 0,
\end{equation}
where violation corresponds to $h_{\text{soft}_i}(x) < 0$.

\subsection{Constraint-Aware Skill Execution}

For each skill, given an associated set of hard and soft constraints,
execution is regulated using an operational-space control barrier
function (OSCBF) filter~\cite{morton2025oscbf}. The filter strictly
enforces hard constraints while biasing execution toward satisfying
soft constraints whenever feasible. This allows the robot to adapt
skill execution to the current context while preserving task intent
and guaranteeing safety. We model the robot with control-affine dynamics in joint space
\begin{equation}
\dot{q} = u,
\end{equation}
where $q$ denotes joint configuration and $u$ is the commanded joint
velocity. Constraints are defined in task space using geometric
quantities extracted from the scene graph. Let
\begin{equation}
    x = FK(q)
\end{equation}
denote the task-space state (e.g., end-effector pose, relative
distance to a human, or orientation), obtained through forward
kinematics. Each constraint is represented as a control barrier
function $h(x) \ge 0$, where $h(x) < 0$ corresponds to unsafe or
undesirable states. Using the chain rule,
\begin{equation}
    \dot{h}(x,u)
=
\nabla h(x)\, J(q)\, u,
\end{equation}
where $J(q)$ is the task Jacobian mapping joint velocities to
task-space velocities. Enforcing
\begin{equation}
    \dot{h}(x,u) \ge -\alpha(h(x))
\end{equation}
guarantees forward invariance of the safe set, where $\alpha$
is an extended class $\mathcal{K}_\infty$ function.

The certified control $u_{\text{cert}}$ is obtained by solving a quadratic program. The objective encourages tracking of the nominal command
$u_{\text{cmd}}$ produced by the primitive skill. Slack variables
$s_i$ allow soft constraints to be relaxed when necessary while
penalizing violations:
\begin{equation}
\begin{aligned}
u_{\text{cert}} =
\arg\min_{u \in \mathcal{U},\, s_i \ge 0}
&\quad
\|u - u_{\text{cmd}}\|_2^2
+ \sum_i s_i^2 \\
\text{s.t.}\quad
& \nabla h_i^{\text{soft}} J u
\ge
-\alpha(h_i^{\text{soft}}) - s_i,
&& \forall i \\
& \nabla h_j^{\text{hard}} J u
\ge
-\alpha(h_j^{\text{hard}}),
&& \forall j 
\\
& \|u - u_{\text{cmd}}\|_2
\le \delta_u.
\end{aligned}
\end{equation}

Here $\delta_u$ defines the maximum allowable deviation from the nominal control command. The resulting controller acts as a safety filter that modifies each primitive skill only as much as necessary to satisfy environment-embodiment constraints. This guarantees certified safety while preserving the structure and intent of the planned skill sequence, enabling context-aware execution without retraining or task-specific controller design.

\section{Simulation Experiments}\label{sec:SimExperiments}
In this section, we address the following research questions through extensive simulation experiments, which provide a safe and scalable evaluation environment: 
\begin{itemize}
    \item[\textbf{Q1.}] Can our framework identify environment-embodiment interaction constraints for caregiving tasks?
    \item[\textbf{Q2.}] Can our system adapt caregiving primitive skills across different environments and robot embodiments based on the environment-embodiment interaction context?
\end{itemize}

\subsection{Simulation Experimental Setup}

We evaluate our framework across diverse robot embodiments,
household environments, and caregiving tasks.
We consider five different robot embodiments: a Kinova Gen3 (7-DoF),
a Kinova 6-DoF arm, a Franka Emika Panda, a UR5, and a WidowX 250. For environments, we evaluate in 130 simulated household~\cite{Matterport3D, ramakrishnan2021hm3d} settings with
varying furniture layouts and object types, including 30 assistive homes~\cite{RCareWorld}.
We assess performance across multiple activities of daily living (ADLs),
including feeding, bathing, grooming, and meal preparation.
For each scenario, ground-truth constraint annotations are labeled by three third-party human annotators.


\textbf{Baselines.}
We evaluate both internal ablations and external baselines. We include ablations: (i) \textbf{Ours w/ GT perception}, where perception modules are replaced with ground-truth state, establishing an upper bound on performance under perfect scene understanding;
(ii) \textbf{Ours w/o dynamic scene graph}, which removes the unified graph representation, representing embodiment and environments with natural language;
(iii) \textbf{Ours w/o constraints}, where fixed primitive skill templates are executed without adapting to constraints. 

We compare against methods that are closely related to our work and explicitly model the environment to ensure safety. Since learning-based models for physical caregiving robots are not yet sufficiently mature due to limited data availability (\secref{sec:intro}), we focus our comparisons on model-based, safety-constrained state-of-the-art methods:
(i) \textbf{SemanticSafe}~\cite{semantic-manipulation}, which represents the environment as an unstructured object list and encodes all the constraints as barrier functions;
(ii) \textbf{Nominal CBF}~\cite{morton2025oscbf}, which enforces a fixed set of pre-specified safety constraints without modeling environment structure. 

\textbf{Metrics.}
We evaluate two primary metrics, following prior work~\cite{morton2025oscbf, semantic-manipulation}:
(1) \emph{Task success rate}, defined as the percentage of episodes in
which the caregiving task is completed without collision;
and (2) \emph{Constraint satisfaction rate}, defined as the fraction of timesteps in which all constraints are satisfied.

\subsection{Evaluating Constraint Synthesis}

We first evaluate whether our framework can correctly identify environment-embodiment interaction constraints given the context. To isolate this component, we provide ground-truth dynamic scene graphs and evaluate the synthesized constraints. Table~\ref{table:simresults_constraintssynthesis} reports the precision and recall of valid constraint synthesis across four ADL tasks. Our system achieves consistently high recall (around 90\%) and strong precision across tasks, indicating that the framework reliably identifies environment-embodiment constraints. For example, in a meal-preparation task, the robot imposes orientation constraints on the knife’s pose during cutting when a human is nearby. Tasks involving more complex spatial interactions, such as grooming, exhibit slightly lower precision. Overall, these results demonstrate that the unified environment-embodiment representation enables reliable constraint generation across diverse contexts.

\begin{table}[ht]
\centering
\caption{\small Constraint Synthesis Precision and Recall on 4 ADLs across Household Environments.}
\begin{tabular}{c cc}
\toprule
   Task       & Precision  & Recall  \\ \hline
Feeding    &    91.0\% $\pm$ 5.1\% &   94.8\% $\pm$ 5.1\%  \\ \hline
Meal Prep. &    88.2\% $\pm$ 10.9\%&   92.1\% $\pm$ 9.5\%  \\ \hline
Grooming   &    86.0\% $\pm$ 8.7\% &   89.0\% $\pm$ 9.5\%  \\ \hline
Bathing    &   90.3\%  $\pm$ 5.8\%&    90.0\% $\pm$ 10.9\% \\
\bottomrule
\end{tabular}
\label{table:simresults_constraintssynthesis}
\end{table}

\begin{table*}[t]
\centering
\caption{Constraint satisfaction rate of the Kinova Gen3 robot performing four ADLs across diverse scenarios.}
\resizebox{\linewidth}{!}{
\begin{tabular}{llccccccc}
\toprule
\textbf{Task} & \textbf{Context (C.)}
& \textbf{Ours w/ GT Perception}
& \textbf{Ours}
& \textbf{Ours w/o graph} 
& \textbf{Ours w/o constraints} 
& \textbf{SemanticsSafe~\cite{semantic-manipulation}} 
& \textbf{Nominal CBF~\cite{morton2025oscbf}} 
\\
\midrule

\multirow{2}{*}{Feeding}
& C1. Dining while the care recipient is watching television
& 100.0\% $\pm$ 0.0\% 
& \textbf{96.1\% $\pm$ 6.3\%} 
& 68.3\% $\pm$ 7.5\% 
& 34.8\% $\pm$ 14.2\% 
& 54.3\% $\pm$ 12.7\% 
& 42.3\% $\pm$ 23.1\% 
\\ 
& C2. Social dining with others
& 95.2\% $\pm$ 9.7\%
& \textbf{90.0\% $\pm$ 14.1\%} 
& 54.3\% $\pm$ 19.7\% 
& 35.8\% $\pm$ 28.3\% 
& 46.1\% $\pm$ 25.9\%
& 42.1\% $\pm$ 28.9\% \\
\midrule

\multirow{2}{*}{Meal Preparation}
& C3. Cooking near the care recipient
& 100.0\% $\pm$ 0.0\%
& \textbf{92.0\% $\pm$ 3.4\%} 
& 77.0\% $\pm$ 18.4\% 
& 45.8\% $\pm$ 24.4\% 
& 72.0\% $\pm$ 26.3\% 
& 69.2\% $\pm$ 29.5\%  \\
& C4. Cooking with a caregiver present
& 89.1\% $\pm$ 8.4\%
& \textbf{82.8\% $\pm$ 12.7\%}
& 67.1\% $\pm$ 17.8\%
& 30.7\% $\pm$ 10.5\%  
& 59.4\% $\pm$ 12.8\%
& 45.8\% $\pm$ 23.4\% \\
\midrule 

\multirow{2}{*}{Bathing}
& C5. Bathtub chair bathing
& 83.1\% $\pm$ 7.8\%
& \textbf{75.2\% $\pm$ 11.9\%} 
& 56.3\% $\pm$ 27.2\% 
& 33.6\% $\pm$ 24.2\% 
& 55.8\% $\pm$ 12.4\% 
& 54.6\% $\pm$ 13.7\% \\
& C6. Bed bathing
& 73.5\% $\pm$ 13.2\%
& \textbf{65.9\% $\pm$ 12.7\%} 
& 45.2\% $\pm$ 19.9\% 
& 41.6\% $\pm$ 17.3\% 
& 42.5\% $\pm$ 18.1\% 
& 43.7\% $\pm$ 16.9\% \\
\midrule

\multirow{2}{*}{Grooming}
& C7. Combing while the care recipient is watching television
& 96.6\% $\pm$ 5.3\%
& \textbf{87.0\% $\pm$ 7.7\%}
& 64.1\% $\pm$ 19.6\%
& 38.4\% $\pm$ 16.4\% 
& 65.4\% $\pm$ 14.6\% 
& 59.8\% $\pm$ 13.4\% \\
& C8. Combing while the care recipient is talking to the caregiver
& 95.7\% $\pm$ 4.6\%
& \textbf{85.3\% $\pm$ 8.9\% }
& 45.7\% $\pm$ 29.1\% 
& 33.3\% $\pm$ 18.1\%
& 55.3\% $\pm$ 26.9\%
& 47.3\% $\pm$ 25.0\%\\
\bottomrule
\end{tabular}}
\label{tab:constraint_violation_percentage}
\end{table*}


\begin{figure}[ht]
\centering
 \includegraphics[width=\linewidth]{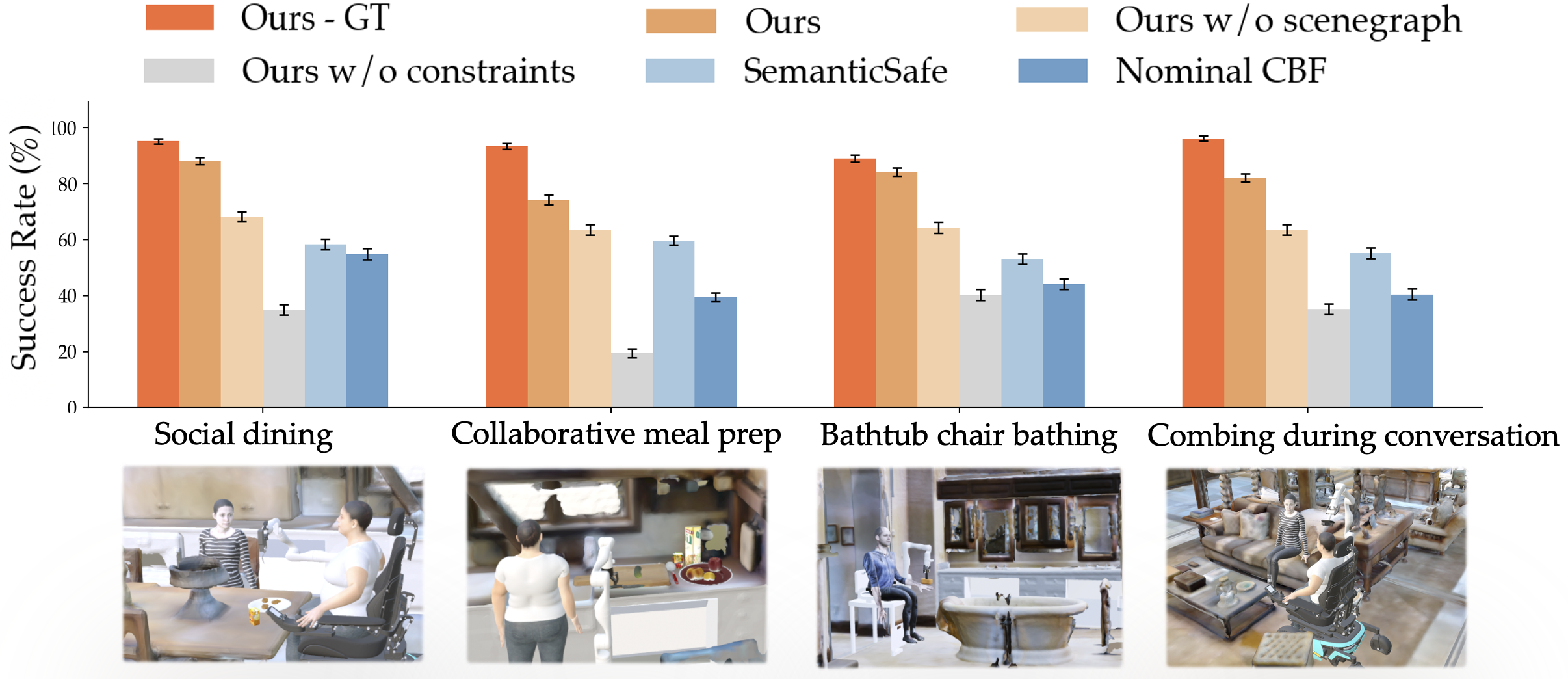}
\caption{Success rate across different scenarios (Table~\ref{tab:constraint_violation_percentage}) on the Kinova Gen3 robot in simulation (More in Appendix).}
\label{fig:sim_sr_across_context}
\vspace{-10pt}
\end{figure}


\subsection{Evaluating Adaptation to Environments}
To study how well our system adapts to diverse environments, we evaluate it across 130 household environments, including 30 assistive homes equipped with assistive devices such as hospital beds and Hoyer lifts, which introduce additional environmental complexity. Table~\ref{tab:constraint_violation_percentage} and Figure~\ref{fig:sim_sr_across_context} report the success rate and average constraint satisfaction rate across eight scenarios spanning feeding, meal preparation, bathing, and grooming tasks with various human activities, e.g., solitary feeding or social dining (scenario details are shown in Table~\ref{tab:constraint_violation_percentage}). Our method consistently achieves the highest success rate (around 95\%) and the highest constraint satisfaction rate (80\%) across all scenarios, closely approaching the upper bound achieved by Ours w/ GT perception, demonstrating our robustness to imperfect perception. 

Our method maintains high reliability, achieving between 82.8\% and 96.1\% constraint satisfaction rate, with an average gap of less than 6\% from the ground-truth perception upper bound. Removing the shared scene graph leads to a substantial degradation in performance. Ours w/o dynamic scene graph drops by about 15 percentage points in success rate across most scenarios. Without explicitly modeling environment-embodiment interaction, the system frequently ignores how the held tool affects the surrounding environment. For example, the robot may point a knife blade toward a human in a social dining context (C2) or move a wet sponge above an electrical device during bathing (C5). Ours w/o constraints consistently performs the worst among ablations, often falling below 45\% in complex interaction scenarios. It ignores human presence in the environment and often causes the robot arm to move dangerously close to, or even collide with, a person. This confirms that fixed skill templates alone are insufficient for safe adaptation to diverse environments.

Compared to the baselines, our method shows a clear advantage. SemanticSafe and Nominal CBF achieve moderate constraint satisfaction rate but remain 15--40 percentage points below our full system in most scenarios. These methods lack a structured representation of the environment. As a result, SemanticSafe often hallucinates constraints between objects and ignores human presence. For example, it may generate a constraint to regulate knife orientation but fail to account for human position, causing the blade to point toward a person. Without explicit modeling of interaction context, the system often hallucinates multiple constraints that make the problem infeasible under the resulting control barriers and frequently stops at a hallucinated barrier. 

Overall, the simulation results show that our framework can reliably synthesize environment-embodiment interaction constraints and adapt caregiving skills across environmental variation, achieving both high task success and a high constraint satisfaction rate.

\subsection{Evaluating Adaptation to Embodiments}

\begin{figure}[t]
\centering
 \includegraphics[width=\linewidth]{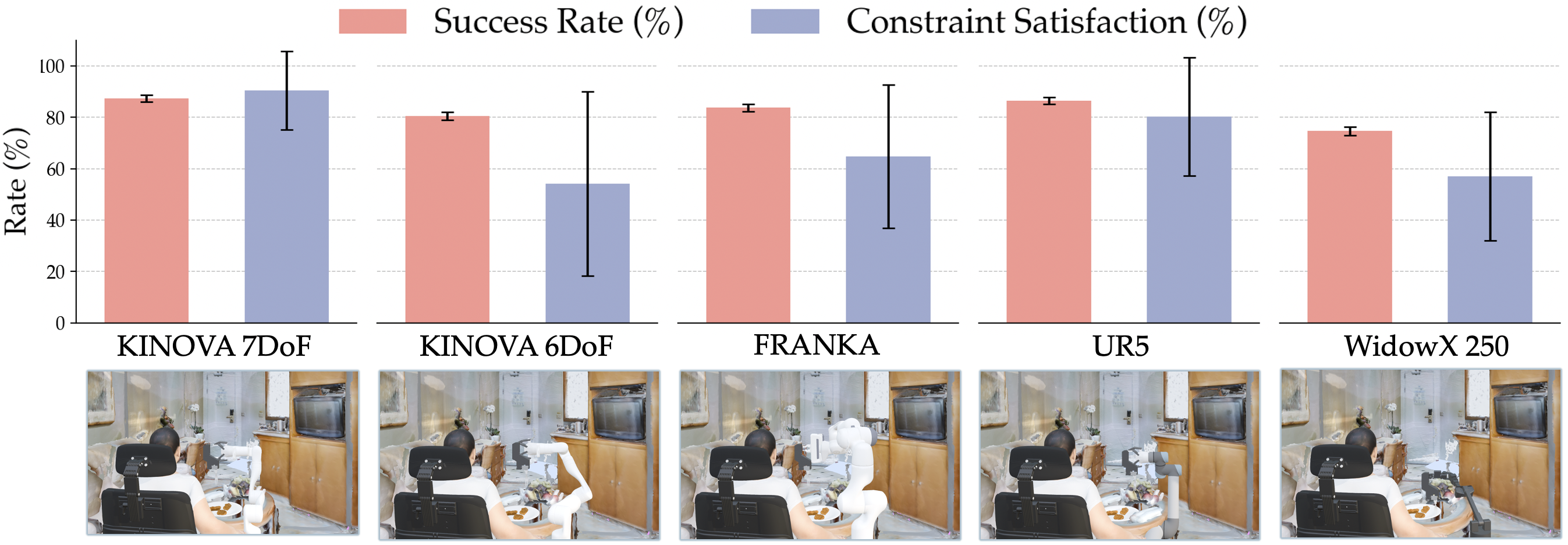}
\caption{Success rate and constraint satisfaction rate of ADL tasks on different robots. 
}
\label{fig:sim_sr_across_robots}
\vspace{-10pt}
\end{figure}

Figure~\ref{fig:sim_sr_across_robots} evaluates our method on different robot embodiments. Our framework maintains strong performance across all platforms, demonstrating robustness to embodiment variations. The Kinova Gen3 achieves the highest performance, with approximately 87\% success rate and 90\% constraint satisfaction rate. The UR5 achieves comparable results (86\% success rate and 80\% constraint satisfaction rate), suggesting that the synthesized constraints generalize well across industrial manipulators with different link geometries. Overall, performance remains high across embodiments despite substantial differences in morphology.

Even on more constrained embodiments such as the WidowX250 and Kinova 6-DoF, which have more restrictive kinematic limits, the system maintains about 75\% success and 55\% constraint satisfaction rate. The performance drop is expected due to embodiment limitations: the controller continues to enforce constraints whenever feasible, but certain constraints become physically difficult to satisfy. For example, the Franka robot’s larger physical footprint makes it harder to avoid blocking the user’s view during feeding. Importantly, the relatively small performance gap between embodiments highlights a key property of our approach: skills represented as interaction templates adapt through embodiment-environment-aware constraint synthesis. These results support our central claim that explicitly modeling embodiment-environment interaction enables adaptive caregiving behaviors.

\section{Real-world User Study}
In this section, we further evaluate our approach on real robotic systems via a Cornell IRB-approved real-world user study measuring perceived safety, behavioral appropriateness, and user satisfaction. Specifically, we address the following research question: Do users perceive that the robot adapts its behavior to the environmental context, and do they feel safe during task execution?

\subsection{Real-World Experimental Setup}
\textbf{Hardware.} We evaluate our framework on two robot embodiments: (i) a Kinova Gen3 7-DoF robotic arm and (ii) a Franka Emika Panda 7-DoF robotic arm. Each robot is equipped with an Intel RealSense D435i RGB-D camera mounted on the wrist for egocentric visual perception. In addition, we use an external Intel RealSense D435i camera to provide a third-person global view. We incorporate force feedback
similar to prior work on physical caregiving systems~\cite{jenamani2025feast}. In contact-rich scenarios (e.g., bite transfer
during feeding), we switch to a task-space compliant controller that uses force feedback to regulate interaction. We also enforce force thresholds for safety; when measured forces exceed predefined limits, the system triggers an emergency stop.

\textbf{Evaluation Scenarios.} We evaluate our approach on two real-world activities of daily living (ADLs): feeding and meal preparation. For feeding, we use a parameterized skill library in~\cite{jenamani2025feast}. For meal preparation, we collected 300 demonstrations of cutting and peeling and fine-tuned Pi0.5 to obtain Vision-Language-Action (VLA)-based primitive skills for these behaviors. Our framework supports both learned and parameterized skills and is agnostic to how skills are obtained. We conduct a user study with five participants per scenario, evaluating both tasks across varied environments and two robot embodiments. For safety reasons, we exclude the no-constraint baseline from the meal-preparation experiments involving sharp tools.

\textbf{Baselines.} We compare two methods in the user study: (i) our full system and (ii) our system without constraints synthesis and satisfaction. 

\textbf{User Study.} We recruit five participants per scenario (3 female, 2 male; ages 20-26). Participants completed consent and demographic forms and received standardized instructions prior to the study. Participants then physically interacted with the robot in each condition in a counterbalanced order and completed a post-task evaluation. Participants evaluated each interaction using a 7-point Likert scale across 3 criteria: perceived safety, behavioral appropriateness, and user satisfaction. 



\label{sec:Setup}

\subsection{Results}
\label{sec:realworldResults}
\begin{figure}[t]
\centering
 \includegraphics[width=\linewidth]{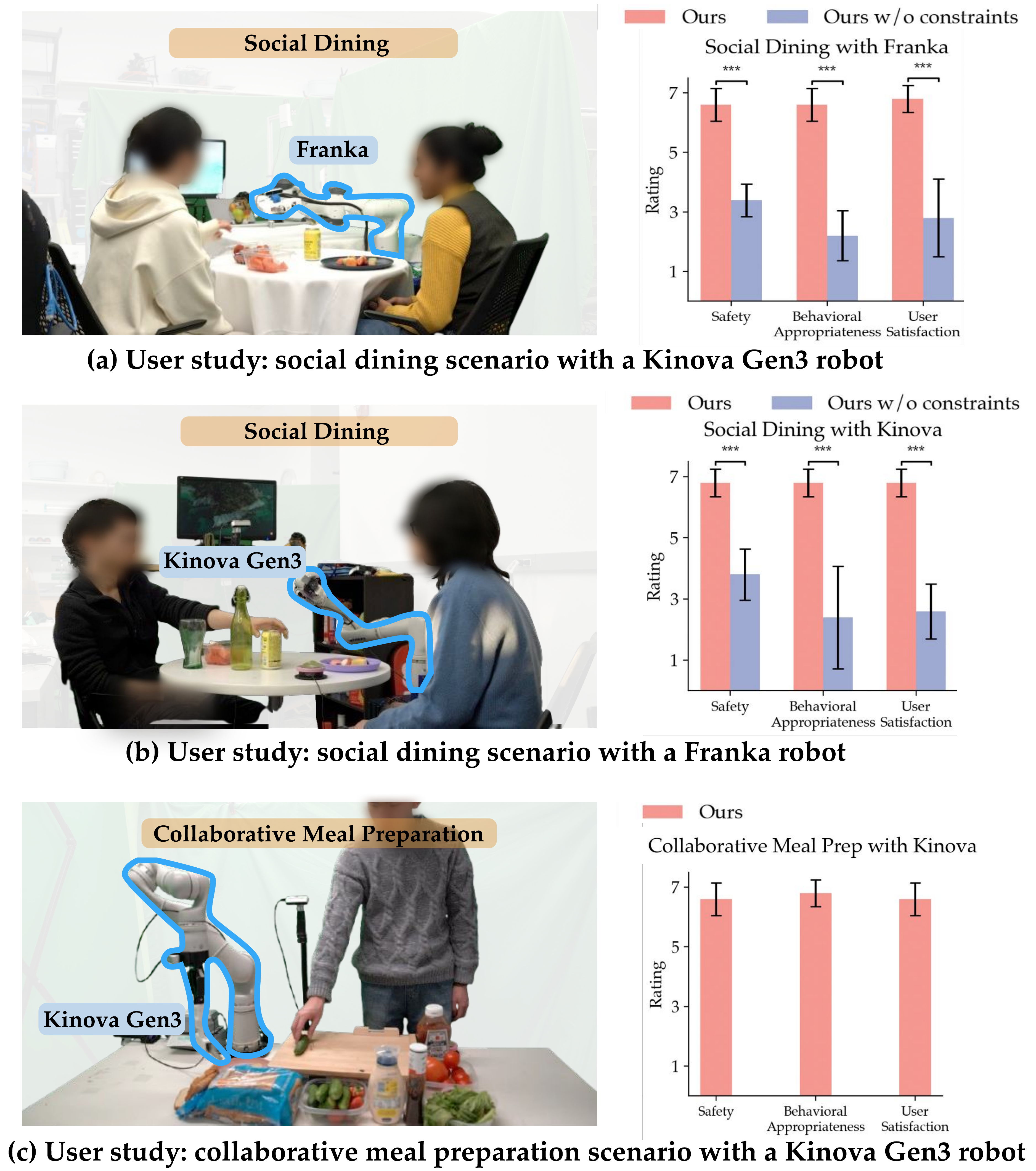}
\caption{User study across diverse environments with Franka and kinova robots in social dining and collaborative cooking.}
\label{fig:realworld_userstudy}
\vspace{-10pt}
\end{figure}

\begin{figure}[ht]
\centering
 \includegraphics[width=\linewidth]{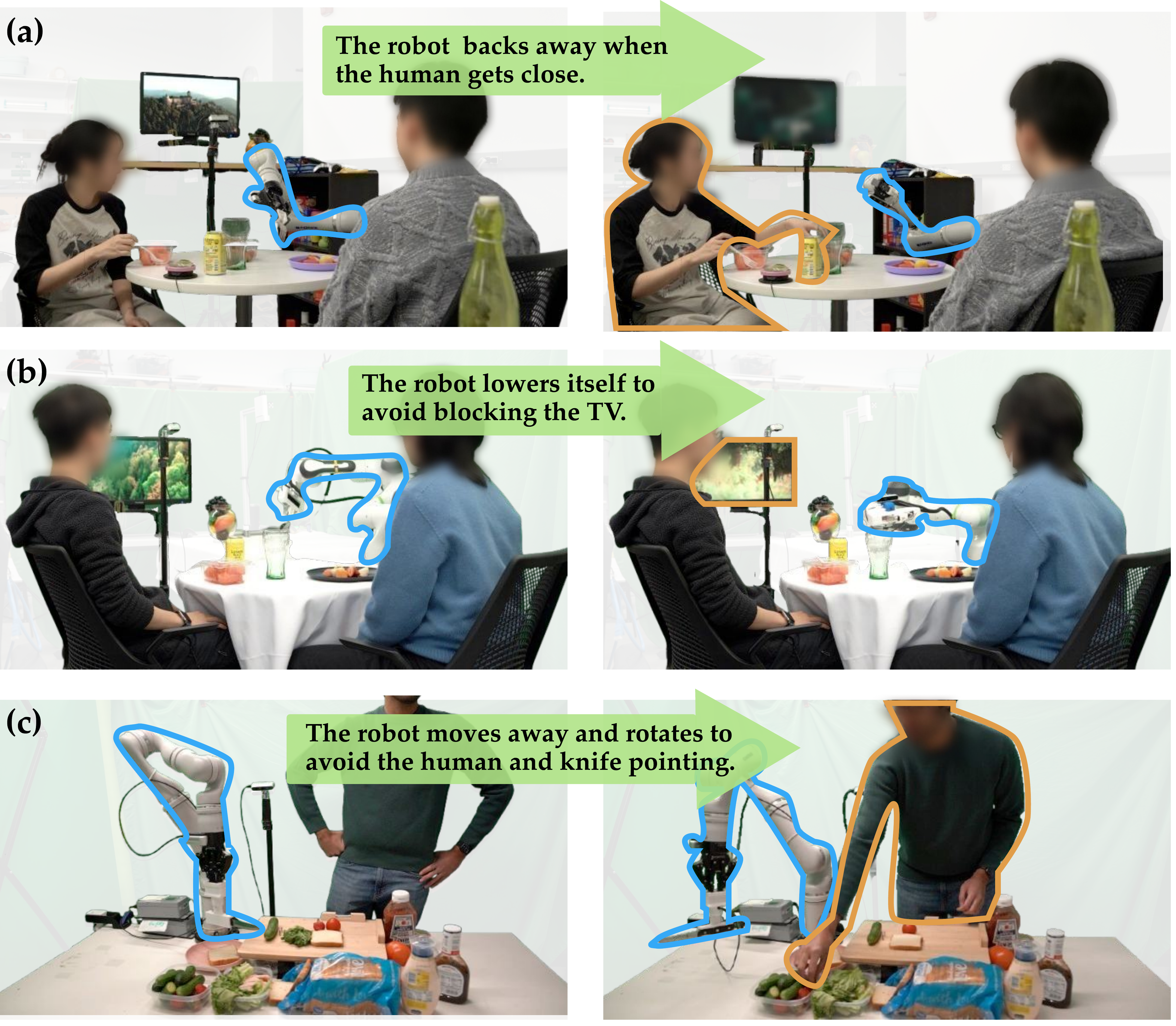}
\caption{Qualitative results. In (a), while the robot is feeding a human in a social dining scenario, another person approaches the table to get food, and the robot backs away to maintain a safe distance. In (b), the robot lowers its motion to avoid blocking the user’s view and then moves away to maintain a safe distance when both attempt to reach the same food. In (c), when a human approaches a robot holding a knife, the robot immediately increases the separation distance and rotates the knife blade away for safety. Once the human leaves the workspace, the robot resumes the task.}
\label{fig:qual_results}
\vspace{-10pt}
\end{figure}

 Results are shown in Figure~\ref{fig:realworld_userstudy}. In social dining scenarios with both the Kinova and Franka robots, our method significantly outperforms the baseline across all three metrics. The largest improvement appears in behavioral appropriateness, suggesting that our framework successfully enables the robot to adapt its behavior to the current context and demonstrating robustness across embodiments. In the collaborative cooking scenario, our method maintains high ratings across all dimensions. The robot explicitly adapts its behavior based on environment-embodiment interaction. For example, when the robot is holding a sharp knife and a human is approaching, it rotates the blade away from the human and maintains a safe distance (shown in Fig.~\ref{fig:qual_results}); in contrast, when the robot is equipped with a standard gripper or the human is far away, no specific orientation constraint is imposed. Participants consistently perceived the robot as safe and appropriate in context, even when the robot was holding a sharp knife. One participant remarked: “What made the robot feel safe was not just that it avoided me, but that it seemed aware of my movement and adjusted the knife orientation in response.”

\subsection{Runtime analysis}
We report runtime across 20 trials on a workstation (RTX 4070 GPU; Intel i7-13700 CPU; 32GB RAM): 
\textit{(i) Perception}: Scene graph construction via open-world detection takes $1.15\pm0.39$s (once per skill). A lightweight human detector runs in the control loop at $(3.85\pm0.14)\times10^{-3}$s for rapid response to human motion. 
\textit{(ii) Planning}: The LLM prunes irrelevant nodes in $1.94\pm0.36$s, and then the PDDL planner generates the skill sequence in $(1.5\pm0.2)\times10^{-4}$s. Constraint synthesis takes $1.96\pm0.52$s and is performed only once per skill (not per timestep). 
\textit{(iii) Control}: The CBF-based QP filter has negligible overhead $(1.5\pm0.3)\times10^{-4}$s, remaining under $3\times10^{-4}$s even with 1000 constraints, ensuring real-time safety enforcement. 
When a human suddenly moves, our method detects the motion in about 0.0004s and reacts within 0.0002s during execution.

\section{Discussions and Limitations} 
\label{sec:discussions}

Our results show that context-aware caregiving behavior emerges from the joint interaction between the environment and the robot embodiment, rather than from either in isolation. In human-centered settings, executing a nominally correct skill without accounting for this interaction can lead to unsafe or socially inappropriate behavior when humans, robots, and surrounding objects co-exist. By explicitly synthesizing environment-embodiment interaction constraints grounded in a shared scene graph, we enable the same skill to be executed differently depending on context, resulting in higher constraint satisfaction in simulation and improved perceived safety and behavioral appropriateness in user studies. Our framework supports both learned and parameterized skills and is agnostic to how skills are obtained. We include a VLA policy to demonstrate compatibility with learned skills, while parameterized skills from existing systems can be integrated without additional data collection. Once a skill is available, it can be reused zero-shot across diverse environments and robot embodiments.

Despite these benefits, the framework has several limitations. First, constraint synthesis relies on prior knowledge encoded in vision-language models (VLMs), and the quality of the generated constraints is therefore tied to the VLM’s understanding of the context; misinterpretations or omissions may lead to missing or overly conservative constraints. 
Second, our framework depends on the quality of perception and scene representation. We provide controller-level safety, but end-to-end safety is conditional on perception accuracy and propagated module errors. Errors in object detection, semantic labeling, or dynamic scene graph construction can propagate to constraint synthesis and affect execution. 
Finally, the scope of adaptation is limited by the predefined primitive skill library. While treating skills as adaptable interaction templates enables reuse across environments and robot embodiments, behaviors outside the coverage of available skills cannot be executed. Supporting online skill discovery or learning new primitives would further improve flexibility and task coverage. 

Our framework currently focuses on safety constraints that must always hold throughout skill execution. One promising direction for future work is to extend the framework to support more expressive temporal constraints. For example, incorporating richer temporal specifications from formal methods such as Signal Temporal Logic (STL), including operators such as \textit{Eventually} and \textit{Until}, would enable reasoning about more complex temporal behaviors and further improve the expressiveness of the framework.

\section*{Acknowledgments}
This work was partly funded by National Science Foundation IIS \#2132846, and CAREER \#2238792. This research was also funded, in part, by the Advanced Research Projects Agency for Health (ARPA-H) Agreement No. 140D042590012. The views and conclusions contained in this document are those of the authors and should not be interpreted as representing the official policies, either expressed or implied, of the U.S. Government.


\bibliographystyle{plainnat}
\bibliography{references}


\end{document}